\documentclass[onecolumn, journal]{IEEEtran}
\ifCLASSINFOpdf
\else
\fi

\usepackage{bm}
\usepackage{CJK}
\usepackage{indentfirst}
\usepackage{amsmath}
\usepackage{multirow}
\usepackage{threeparttable}
\usepackage{tabularx}
\usepackage{graphicx}
\usepackage{subfigure}
\usepackage{float}
\usepackage{color}
\usepackage{booktabs}
\usepackage{amsmath,amssymb}
\usepackage{epstopdf}
\usepackage{inputenc}
\usepackage{diagbox}
\usepackage{svg}
\usepackage{hyperref}

\makeatletter
\newif\if@restonecol
\makeatother

\usepackage[linesnumbered,ruled,vlined]{algorithm2e}   
\usepackage{algpseudocode}
\usepackage{amsmath}

\begin{document}
\title{Spatiotemporal Capsule Neural \\ Network for Vehicle Trajectory Prediction}
\author{
		Yan~Qin,~\IEEEmembership{Member, IEEE,}
			Yong Liang~Guan,~\IEEEmembership{Senior Member, IEEE,}
             	and Chau~Yuen,~\IEEEmembership{Fellow,~IEEE}
\thanks{This research is supported by A*STAR under its RIE2020 Advanced Manufacturing and Engineering (AME) Industry Alignment Fund-Pre Positioning (IAF-PP) (Grant No. A19D6a0053). Any opinions, findings and conclusions or recommendations expressed in this material are those of the author(s) and do not reflect the views of A*STAR. \textit{(Corresponding author: Chau Yuen.)}}
\thanks{Y. Qin, Y. L. Guan, and C. Yuen are with the School of Electrical and Electronic Engineering, Nanyang Technological University, 639798 Singapore. (e-mail: yan.qin@ntu.edu.sg, eylguan@ntu.edu.sg, chau.yuen@ntu.edu.sg)}
\thanks{Copyright (c) 2015 IEEE. Personal use of this material is permitted. However, permission to use this material for any other purposes must be obtained from the IEEE by sending a request to pubs-permissions@ieee.org.}
}

%

\markboth{}
{Shell \MakeLowercase{\textit{et al.}}: Bare Demo of IEEEtran.cls for IEEE Journals}
%


\maketitle
\begin{abstract}
Through advancement of the Vehicle-to-Everything (V2X) network, road safety, energy consumption, and traffic efficiency can be significantly improved. An accurate vehicle trajectory prediction benefits communication traffic management and network resource allocation for the real-time application of the V2X network. Recurrent neural networks and their variants have been reported in recent research to predict vehicle mobility. However, the spatial attribute of vehicle movement behavior has been overlooked, resulting in incomplete information utilization. To bridge this gap, we put forward for the first time a hierarchical trajectory prediction structure using the capsule neural network (CapsNet) with three sequential components. First, the geographic information is transformed into a grid map presentation, describing vehicle mobility distribution spatially and temporally. Second, CapsNet serves as the core model to embed local temporal and global spatial correlation through hierarchical capsules. Finally, extensive experiments conducted on actual taxi mobility data collected in Porto city (Portugal) and Singapore show that the proposed method outperforms the state-of-the-art methods.
\end{abstract}

\begin{IEEEkeywords}
Vehicle-to-everything network, Next location prediction, Machine learning, Capsule neural network.
\end{IEEEkeywords}

\section{Introduction}
\IEEEPARstart{T}{he} rapid advancements in self-driving vehicles, communication transceivers, and smart sensors speed up the progress of the Vehicle-to-Everything (V2X) network, which has the potential to bring forth a safe, intelligent, and efficient transportation system. The success of V2X highly depends on the quality of the communication network. Therefore, an accurate prediction of the vehicle quantity plays an important role in optimizing the communication network \cite{2022Zhang}-\cite{Ref17}. Imagine that a large number of intelligent vehicles swarmed into business areas during rush hours will overwhelm communication and computation resources in a short time, causing safety incidences for autonomous vehicles when the service quality is undermined. The real-time prediction of vehicle traffic in an area restricted by limited communication and computation resources allows for alleviating the challenges raised by time-varying network demands. Further, real-time vehicle trajectory prediction saves time for network resource allocators to timely respond, inferring the possible network traffic consumption and conducting predictive tasks offloading in advance. Therefore, vehicle trajectory prediction critically supports to ensure a more reliable vehicular network and its associated services, such as vehicle security \cite{RefSecurity1}\cite{RefSecurity2}, self-driving \cite{SelfDrive}, offloading task decisions \cite{2020Offloading}\cite{2021Offloading1}.

Recently, inferring vehicle future locations has been accelerated by the availability of vehicle trajectory information. Nowadays, real-time vehicle mobility is located by in-vehicle Global Positioning System (GPS) receivers and then transmitted through road-side units, base stations, and collaboratively communicated with other vehicles. Due to the diversity of vehicles' regular movements, data-driven vehicle trajectory prediction is particularly appealing since the historical trajectory information helps to sense the future locations \cite{2015Petit}\cite{Ref15}. In light of this, data-driven vehicle trajectory prediction methods have been extensively studied, ranging from the traditional state transition model to deep neural networks. For instance, a low-order Markov-based prediction model has been proposed by Zhu et al. \cite{2014Zhu} to depict vehicle mobility. Further, Irio et al. \cite{2021Irio1} described the hidden states with a fixed-length sequence of consecutive vehicle locations rather than an individual location. Afterward, a hidden Markov model for trajectory prediction has been proposed to capture the sequential nature of individual trajectory. However, Markov model-based approaches suffer from the curse of data dimensionality when a long historical route attempts to be considered for trajectory prediction. This limitation caps the prediction accuracy as a short historical route only contains limited information.

\begin{figure*}[!htb]
\centering
\setlength{\abovecaptionskip}{-0.2cm}
\includegraphics[scale=0.35]{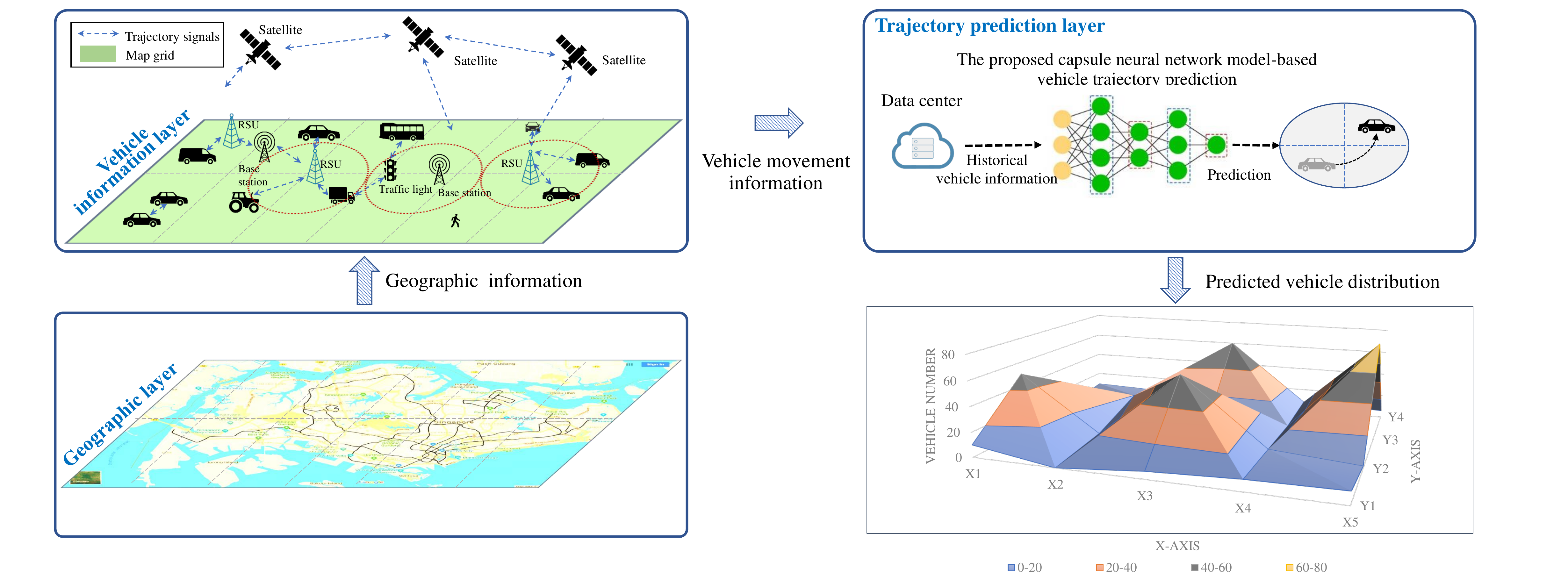}
\caption{Overall framework of vehicle trajectory prediction with three hierarchical layers, namely, the lower geographic layer, the middle vehicle information layer, and the upper trajectory prediction layer.}
\label{FIG1}
\end{figure*}

To completely leverage the long-term dependent correlations within time-series data, recurrent neural network (RNN) and their well-known variants \cite{1997LSTM}\cite{Ref12}, like long short-term memory (LSTM) and gated recurrent unit (GRU), have attracted extensive attention with their superiorities in temporal representation \cite{Ref13new}-\cite{RefZhouqi}. To handle the handover in vehicular networks and vehicle-based sensing data, Liu et al. \cite{2019Liu} applied GRU to predict vehicle mobility with edge devices. Adege et al. \cite{2020Adege} put forward a hybrid approach of principal component analysis and GRU for mobility predictions in a wireless urban area. Irio et al. \cite{2021Irio} compared RNN-based future location prediction with statistical inference-based methods, demonstrating that the RNN approach is competitive in both accuracy and computational time. More recently, Ip et al. \cite{Ip2021} adopted LSTM to predict the future locations of vehicles with several different prior locations information. The results show that the much prior information, the higher the prediction accuracy. Furthermore, Choi et al. \cite{Choi2019} introduced the attention mechanism to pay attention to the traffic flow of adjacent areas spatially. Liu et al. \cite{Liu2022} developed the prediction model considering the uncertainty of motion behavior. Although approaches from the RNN family have achieved remarkable success in sequential learning, they may not be perfectly suitable for trajectory prediction. To implement vehicle trajectory prediction, a large-scale urban area will be split into a series of small-scale grids with a fixed area according to communication distance. Correspondingly, the sequential GPS signal will be transformed into a vector with grid information. The trajectory prediction aims to utilize past grid information to predict future grid locations. LSTM fails to exploit the spatial relationship of vehicle locations during travel history. However, such spatial information is non-negligible. It reasonably expects that a vehicle moves from a specific place towards its surrounding locations rather than a faraway location in a short period. Izquierdo et al. \cite{Izquierdo2022} predicted vehicle trajectories in highway scenarios using efficient bird’s eye view representations and convolutional neural networks (CNN). Although CNN possesses the capability of spatial feature representation, it suffers from information loss caused by pooling operation, which averages the spatial relationships among different parts of an entity \cite{2020SubCapsNet}. To support communication or computation resource allocation in the V2X network, more reliable and accurate vehicle trajectory prediction is required to leave enough time for resource allocators to act when a traffic surge is coming. To achieve this purpose, some challenges need to be further addressed:

\begin{itemize}
\item Although more past travel locations contribute to improving the estimation accuracy of future location, it challenges the privacy-persevering if the drivers' travel history is utilized too much;
\item Changes in spatial relationships are overlooked due to the average or maximum pooling, discounting the efficacy of the joint integration of temporal correlation and spatial information.
\end{itemize}

Nowadays, the capsule neural network (CapsNet) invented by Sabour et al. \cite{Ref11} has been widely used to carry out the equivalent representation of entities in the image processing field. Removing pooling operation \cite{CapsNet1} from the network structure avoids information loss. Moreover, CapsNet introduces a novel concept, i.e., capsule, to describe the local features with vector representation rather than scalar, contributing to embedding position and direction information into local features. As such, CapsNet is more suitable for carrying spatial and temporal information for trajectory prediction. This further motivates us to take advantage of the benefits of CapsNet for trajectory prediction. Furthermore, a general vehicle trajectory prediction framework with three hierarchical layers is proposed, as shown in Fig. \ref{FIG1}. The bottom geographic layer is static and used to indicate the physical space, such as buildings, rivers, roads, etc. In the middle vehicle information layer, it reveals the dynamic and time-varying movements of vehicles, in which raw data is encoded to keep the spatial and temporal information. The upper trajectory prediction layer estimates the next location of each vehicle through the past trajectory and the well-trained CapsNet. Through the second and third layers, the past travel trajectory used for prediction has been reduced to achieve privacy-preserving. Consequently, accurate vehicle trajectory prediction is gained, and vehicle traffic flow information is reached through comprehensive data analysis. Although the core of this work focuses on the prediction accuracy of the vehicle trajectory in the V2X network, it benefits a lot of the downstream topics, one of which is the mobility-aware task-offloading decision. Specifically, the ever-increasing high-traffic and heavy-computation tasks from intelligent vehicles are expected largely in the V2X network. However, dynamic topology changes in vehicle mobility and packet drop make offloading decisions more complex. Therefore, an accurate estimation of the vehicle trajectory prediction is particularly important to sense the amounts of tasks. The major contributions of this work are summarized as follows:

\begin{itemize}
\item The vehicle travel trajectory is processed in a way so as to retain the spatial information, and user privacy-preserving is enhanced through the temporal data construction with a specific length;
\item A CapsNet-based vehicle trajectory model is put forward to jointly learn the spatial and temporal features for the first time;
\item Through the designed three-layer hierarchical architecture with lossless information fusion, extensive experiments and comparisons with benchmark datasets have been conducted to verify the performance of the proposed model, providing dynamic vehicle distribution information to assist network resource allocation.
\end{itemize}



The structure of the remaining parts is organized as follows. Section II describes the data preparation of vehicle trajectory and formulates the prediction problems in V2X scenario. In Section III, we detail the proposed model from data representation with grid map, CapsNet-based prediction model, and online prediction. We present the performance of the proposed method and compare it with existing approaches in Section IV. Finally, Section V concludes this work.

\section{Data Preparation and Problem Formulation}
In the geographic map, a vehicle's movement state at a specific time is located by GPS coordinates $s=(x, y)$, where $x$ denotes the longitude position and $y$ stands for the latitude position. In practice, GPS signals are recorded with in-vehicle devices or mobile phones of drivers. A list of visited locations are recorded sequentially as $\mathbf{S}_k = \{s_1, s_2, \dots, s_k,  \dots\}$. It should be noted that the real starting/ending information of a trajectory may be removed in some datasets \cite{Grab} to protect personal privacy.

\subsection{Data Preparation}
The hardware of V2X network, such as base stations, road side units, edge servers, etc., are deployed away from each other, in charging of communication and computation tasks. The limited number of devices lead to restricted resources in a specific area. As such, tasks load of the V2X network may suffer from violent fluctuations when the incoming and outgoing vehicles of a specific region heavily change. Due to spatial allocation of the network resource, it is more reasonable to separate the entire area into a series of small zones to facilitate vehicle trajectory prediction. Here, grid map representation serves as the crucial data preparation, assigning GPS coordinates into a series of city grids. Specifically, the grid map is constructed by dividing the geographic area of a two-dimensional coordinate space into small grids with fixed width and height. In practice, the size of a grid has a close relationship with the density of communication bases, and road side units, which could be inferred from a priori knowledge \cite{Ref13}\cite{Ref14}.

The maximum and minimum of longitude and latitude in the concerned area are denoted as ($LO_{max}$, $LO_{min}$) and ($LA_{max}$, $LA_{min}$), respectively. The intervals of each grid are specified as $(LO_{max}-LO_{min})/N_o$ and $(LA_{max}-LA_{min})/N_a$, where $N_o$ and $N_a$ are the number of units along the longitude and latitude directions in the grid map. By matching the latitude and longitude information covered by each grid, the vehicle's movement trajectory from one grid to another can be calculated. Assuming the whole area is equally divided into $G$ grids with the same scale, the $v^{th}$ traveling trajectory of a vehicle can be described through a list of visited grids sequentially with $\mathbf r_v = \{r_1^v, r_2^v, ..., r_{N_v}^v \}$, where $G=G_a \cdot G_o$, $r_1^v$ is the starting position, $r_{N_v}^v$ denotes the ending position, $N_v$ stands for the duration of the trajectory, and $r_i^v$ belongs to the collection of $[1, 2, ..., G]$, each of which indicates a specific grid.

\begin{figure}[!htb]
\centering
\includegraphics[scale=0.85]{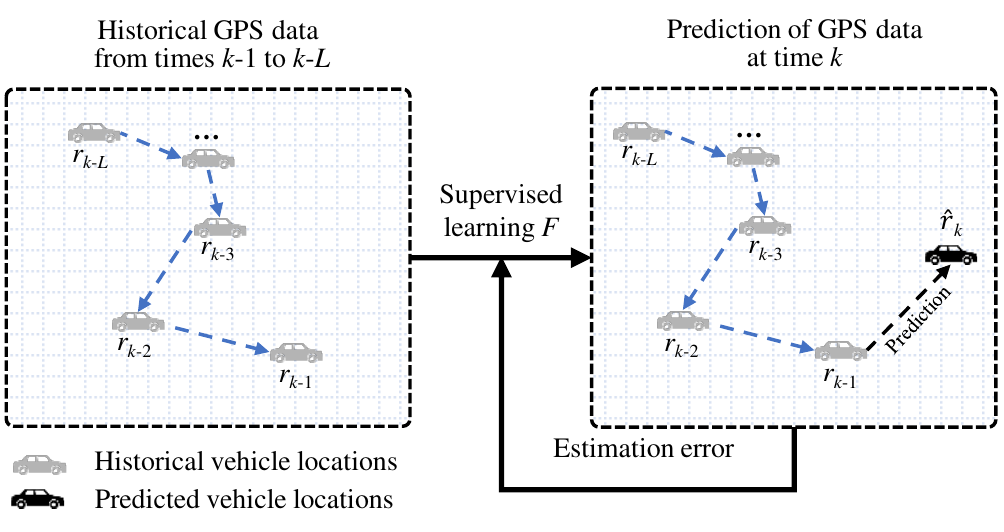}
\caption{Overall structure of trajectory prediction in supervised learning.}
\label{FIG2}
\end{figure}

\subsection{Problem Formulation}
As shown in Fig. \ref{FIG2}, vehicle trajectory prediction is a supervised learning, in which applying classification task is to train a predictor $F$ on a period of historical GPS measurements as labels. As such, a mathematical representation is arranged below,
\begin{equation}
\hat {r}_k = F\{\mathbf {v}_{Hist}(k) |\Theta \}
\end{equation}
where $F(\cdot)$ is the predictor to be trained, $\Theta$ is the parameters of $F(\cdot)$, and $\mathbf {v}_{Hist}(k)=\{r_{k-1}, \dots, r_{k-L}\}$ is the historical measurements with a duration of $L$.

Once the training phase is completed, the well-trained model can be deployed to make predictions on real-time measurements to predict the next location.

\section{Capsule Neural Network-based \\ Vehicle Trajectory Prediction}
\begin{figure*}[!htb]
\centering
\includegraphics[scale=0.55]{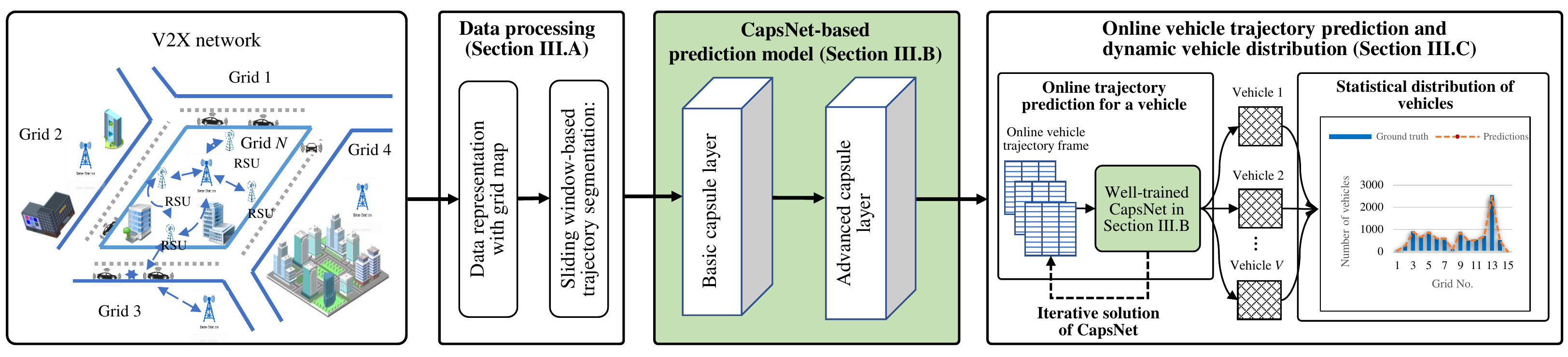}
\caption{The flow chart of the proposed vehicle trajectory prediction using CapsNet.}
\label{FIG3}
\end{figure*}

Fig. \ref{FIG3} specifics three sequential parts to implement the proposed method. First, data processing transforms raw GPS data into geographic information through grid map representation. Second, a CapsNet-based model has been put forward for trajectory prediction. Third, when the model has been well-trained, online trajectory prediction is performed.

\subsection{Data Representation with Grid Map}
\subsubsection{Data Encoding} 
\begin{figure*}[!htb]
\centering
\includegraphics[scale=0.64]{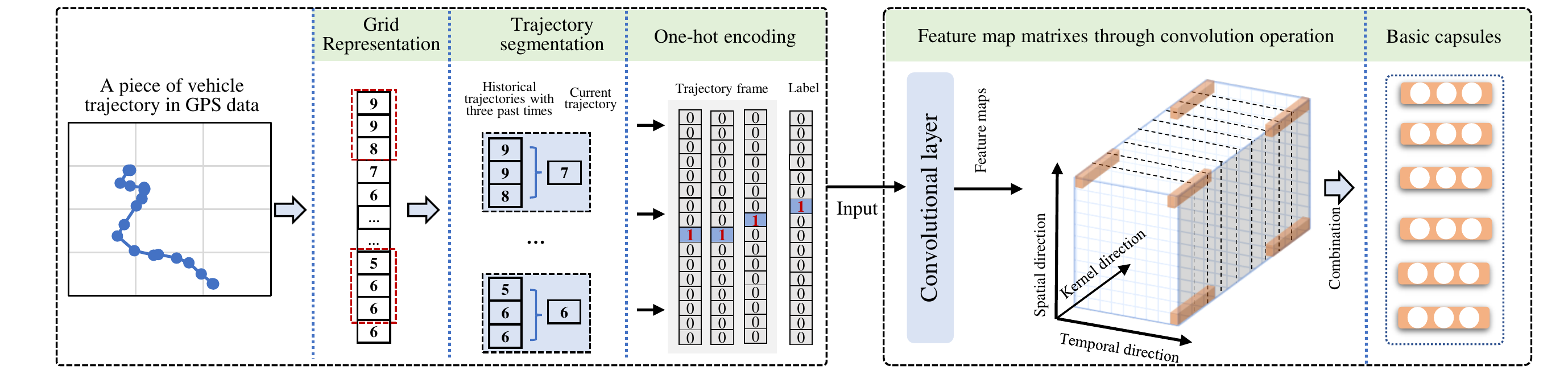}
\caption{The steps to perform sliding window-based trajectory segmentation and generate basic capsules through the convolutional operation.}
\label{FIG4}
\end{figure*}

We process the actual GPS data according to the procedure given in Section II.A. Afterward, to take advantage of capture the spatial information, data encoding is implemented to convert the scalar grid information into a vector grid with binary coding, as shown in Fig. \ref{FIG4}.

\subsubsection{Sliding Window-based Trajectory Segmentation} Sliding window technology is used to divide the entire trajectory ${\mathbf r}_v$ into a series of temporal data units with the same length, as shown in Fig. \ref{FIG4}. We collect $L$ consecutive grids to construct the historical matrix $\mathbf r \in \mathbb{R}^{G \times L}$, named as trajectory frame, to predict the next grid of a vehicle $y_k$ at time $k$. As such, the input data of the estimation model are the collection of trajectory frames, which is denoted as $\mathbf T\in \mathbb{R}^{I \times G \times L}$ and the corresponding output is labeled as $\mathbf Y\in \mathbb{R}^{I \times 1}$, where $I$ is the number of trajectory frames.

\subsection{CapsNet-based Prediction Model}
Capsules are originally defined as features with multiple dimensions in a vector format. Basic and advanced capsules are constructed with low-dimensional and high-dimensional vectors to present local and global feature representation, respectively. Basic capsules are the smallest element of whole feature space, and they are combined to construct advanced capsules with higher dimension and presentation capability.

\subsubsection{Basic Capsule Layer} This layer aims to transform the trajectory frame into vector-based feature outputs with the capsule concept. Compared with the traditional scalar features, the introduction of the capsule contributes to the use of multi-dimensional features to describe the various movements of vehicles. Specifically, the spatial relationship from one grid to another can be easily captured with the convolution operation during the construction of basic capsules.

First, the trajectory frame is operated through typical convolution kernels, generating a series of feature maps. It is worth mentioning that the weights of each kernel are randomly given at first and then will be updated iteratively. Given the $m^{th}$ kernel is ${\mathbf \Phi}_m$, its corresponding feature map could be calculated below,
\begin{equation}
\mathbf F_m  = tanh(\Phi_m \ast \mathbf T_{i} + \mathbf b_m)  \quad m \in [1, M] \\
\end{equation}
where $\mathbf b_m$ is the $m^{th}$ bias term, $\ast$ is the convolutional operation, $M$ is the number of kernels, and $tanh(\cdot)$ is the hyperbolic tangent function.


The obtained feature maps construct a three-dimensional data matrix along the feature map direction, as shown in Fig. \ref{FIG4}. Each feature map is the abstract of the original trajectory frame, which focuses on local details different from each other. Here, the basic capsules are defined as the multiple vectors picking up from the same position along the feature map direction. In this way, the original local features described by feature maps are transformed into the basic capsules. The benefits are twofold. First, each dimension of a basic capsule indicates a specific property for trajectory prediction. Second, the length of a basic capsule reveals the probability of visiting the corresponding grid.

\begin{algorithm}[!htb]
\caption{Iterative training of CapsNet}
\scriptsize
\LinesNumbered
\KwIn{Trajectory frame ${\mathbf T_{i}}$ and the associated ground truth $ y(i)$, the random given convolutional kernel ${\mathbf \Phi}_m$, the initial advanced capsules $\mathbf v_j$, and the weights $\Theta_c$ of fully connected layer;}
\KwOut{The well learned ${\mathbf \Phi}_m$, the weights $\mathbf W_{i,j}$ between basic capsule $i$ and advanced capsule $j$, and weights $\Theta_c$;}
Initialize convolutional kernel $\Phi_{m}$, the prior probability $b_{i, j}$ that basic capsule $i$ coupled to advanced capsule $j$; Number of epochs; \\
\Repeat
{The given epochs are reached or early stopping when no improvements are observed on the validation dataset.}
{
1. Obtain feature maps $\mathbf F_m$ according to Eq. (2);

2. Get a series of basic capsules $\mathbf u_i$ from $\mathbf F_m$ according to Fig. 4;

3. Generate a series of advanced capsules  $\mathbf v_j$;

{\Repeat (//Update weight between basic capsules and advanced capsules)
{Given iterations (here is 3)}
{4. The normalized probability $c_{i, j}$ with consideration of the capsule $i$ to other advanced capsules is calculated as below,
\begin{equation}
c_{i, j}= \frac{\exp (b_{i, j})} {\sum_{k} \exp (b_{i, k})}  \quad \forall i, j
\tag{A1}
\end{equation}

5. By weighting sum over all input vectors $\hat{\mathbf u}_{j \mid i}$ from the basic capsules, the input of an advanced capsule $j$ is calculated below,
\begin{equation}
\mathbf s_{j}=\sum_{i} c_{i, j} \hat{\mathbf u}_{j \mid i}
\tag{A2}
\end{equation}
where $\hat{\mathbf u}_{j \mid i}=\mathbf W_{i, j} \mathbf u_{i}$.

6. The initial coupling coefficients are refined by updating $b_{i, j}$ according to the following rule,
\begin{equation}
b_{i, j}=b_{i, j}+\hat{\mathbf u}_{j \mid i} \cdot \mathbf v_{j}
\tag{A3}
\end{equation}
where $\mathbf v_j$ is the output of the $j$th advanced capsule, which is defined below,
\begin{equation}
\mathbf v_{j}=\frac{\|\mathbf s_{j}\|^{2}}{1+\| \mathbf s_{j}\|^{2}} \frac{\mathbf s_{j}}{\|\mathbf s_{j}\|}
\tag{A4}
\end{equation}
}

7. Regress the advanced capsules with the labels according to Eq. (3);

8. With the estimation error, update the network parameters $\mathbf \Phi_m$ and the fully connected layer $\Theta_c$ using Adam optimization algorithm.
}
}
\end{algorithm}

\subsubsection{Advanced Capsule Layer} Overlapped vehicle movements are frequently observed daily, resulting in repetitive basic capsules to describe the same or similar information. In addition, some basic capsules are insensitive to distinguish different labels as their dimensions are not high enough. As such, the advanced capsules are necessarily defined as high-dimensional vectors to address the above two challenges, capable of describing high-level and non-repetitive features. Advanced capsules are the combinations of local features given by basic capsules. Generally, the dimension of advanced capsules is much higher than that of the basic capsules. We use a fully connected layer with definite neurons to construct the advanced capsules. The adjacent neurons will be assigned to the same group to represent the $j^{th}$ advanced capsule.

\subsubsection{Iterative Solution of CapsNet} CapsNet is trained in an iterative way, which employs the ``dynamic routing" mechanism \cite{Ref11} to update weights and bias of the network. Dynamic routing is mainly used for optimizing parameters linking basic capsules and advanced capsules. The basic capsules that are highly related to the advanced capsules will be retained. Otherwise, the basic capsules will be removed by setting the weight as almost zero. On the basis of this, the short vectors shrink to approximately zero, and long vectors approach a length slightly below one. In this way, the length of the output vector of a capsule stands for the probability of the existence of extracted local features. The specifics of network training are described in Algorithm 1.

Multiple linear regression layers with a proper dropout probability are used for regressing the advanced capsules on labeled data. Using $F(\cdot)$ to stand for the learned relationship, the location of a vehicle at time $k$ can be estimated as below,
\begin{equation}
\hat {y}_k = F(\mathbf {v}_k|\Theta_c)
\end{equation}
where $\mathbf {v}_k$ is the collection of outputs from the advanced capsule layers for the $k^{th}$ data frame, and $\Theta_c$ indicates the collection of model parameters in fully connected layers.

The five hyper-parameters in CapsNet are summarized in Table \ref{Table1}. Among them, two parameters, the number of filters $M$ and the dimension of basic capsules $D$, are the crucial ones. They are determined by sensitivity analysis to find the proper value through the validation data \cite{TemporalCapsNet}. Correspondingly, the other remaining parameters can be determined by $M$ and $D$. Dimensions of the basic capsule are calculated by $M/D$. The advanced capsule's dimension is twice the dimensions of the basic capsules. The number of advanced capsules is the same as the number of grids.

\begin{table}[!ht]
\renewcommand{\arraystretch}{1.2}
\centering
\caption{Summary of the hyper-parameters used in CapsNet.}
\scriptsize
\begin{tabular}{p{2cm} p{3.3cm} p{2.5cm}}
\hline
\hline
\textbf{Type} & \textbf{Parameter name} & \textbf{Recommended value} \\
\hline
\multirow{4}{*}{Independent parameter} & No. of filters ($M$) & Depending on sensitive analysis \\
\cline{2-3}
& Dimension of basic capsules ($D$) & Depending on sensitive analysis \\
\cline{2-3}
\hline
\multirow{3}{*}{Dependent parameter} & No. of advanced capsules & $G$ \\
\cline{2-3}
& Dimension of advanced capsules & $2D$ \\
\cline{2-3}
& Dimension of basic capsules & $M/D$ \\
\hline
\hline
\end{tabular}
\begin{tablenotes}
\item[1] $G$ is the number of divided grids in the urban area.
\end{tablenotes}
\label{Table1}
\end{table}

\begin{figure}[!htb]
\centering
\includegraphics[scale=0.65]{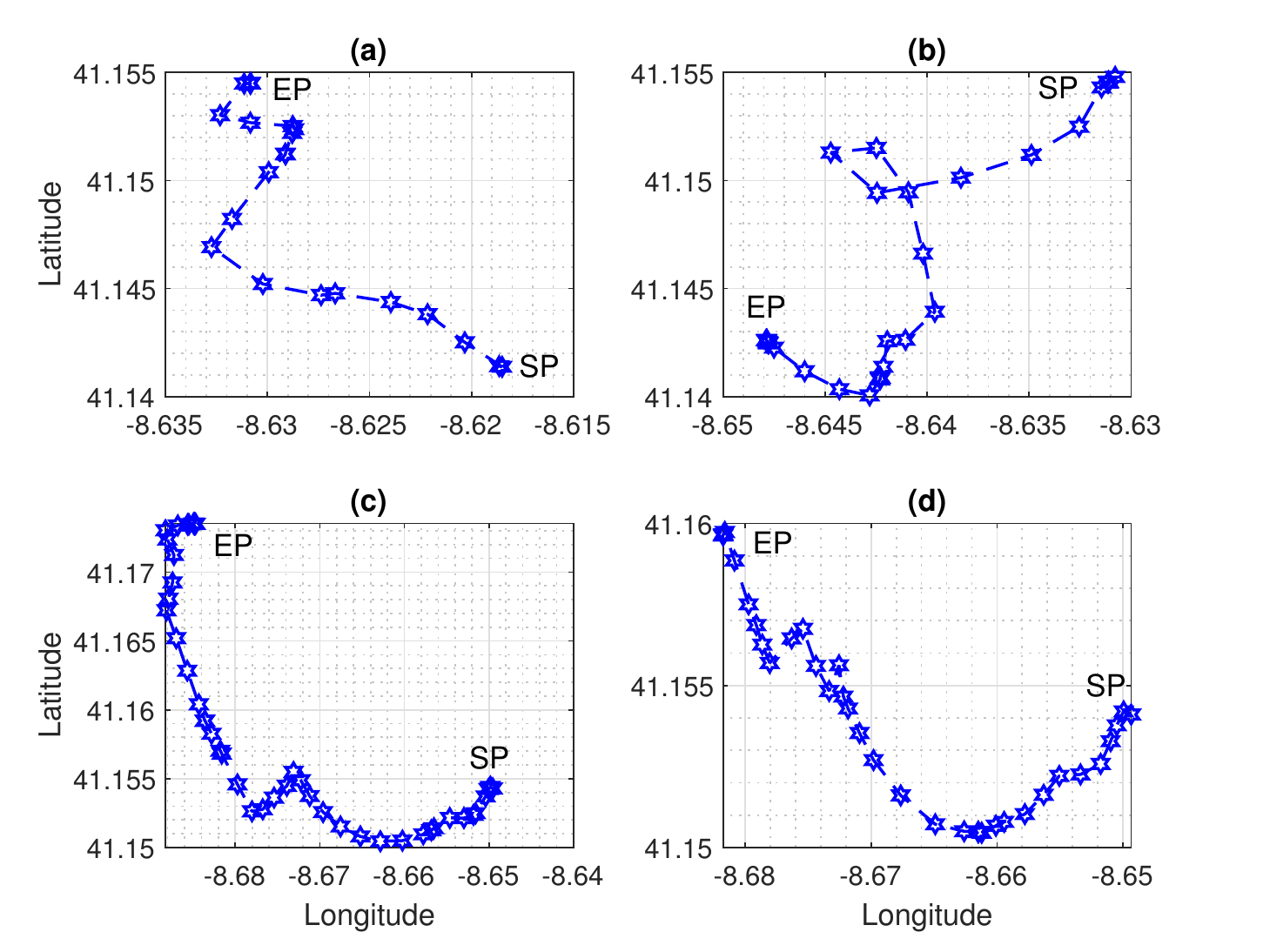}
\caption{Four trips (trajectories) of a vehicle in the dataset: (a) first trajectory, (b) second trajectory, (c) third trajectory, and (d) fourth trajectory (SP indicates the starting position and EP stands for the ending position). }
\label{FIG6}
\end{figure}

\subsection{Online Vehicle Trajectory Prediction and Distribution}
With the well-trained model, online trajectory can be predicted with the available historical trajectory. Afterward, the distribution of all vehicles in city grids can be deduced. The dynamic vehicle distribution with time evolution is available.

\subsubsection{Online Trajectory Prediction for A Testing Vehicle}
\begin{itemize}
\item Step 1: Collect the raw GPS signal of a testing vehicle as $\mathbf r_{Test} \in \mathbb{R}^{L \times 1}$.

\item Step 2: Transform GPS signal of the testing vehicle $\mathbf r_{Test}$ into the grid map representation with grid trajectory according to the ways given in Sections II.A and III.A, which is denoted as $\mathbf g_{Test}\in \mathbb{R}^{G \times L}$.

\item Step 3: Feed $\mathbf g_{Test}$ into the trained CapsNet model given in Section III.B and obtain the advanced capsules $\mathbf v_{Test}$.

\item Step 4: Calculate online prediction results as follows,
\begin{equation}
\hat {y}_k = F(\mathbf {v}_{Test} |\Theta_c)
\end{equation}
where $F(\cdot)$ is the well-trained function from Eq. (3).
\end{itemize}

\begin{figure*}[htb]
\centering

\subfigure[]
{
\begin{minipage}[t]{0.45\linewidth}
\includegraphics[width=6cm]{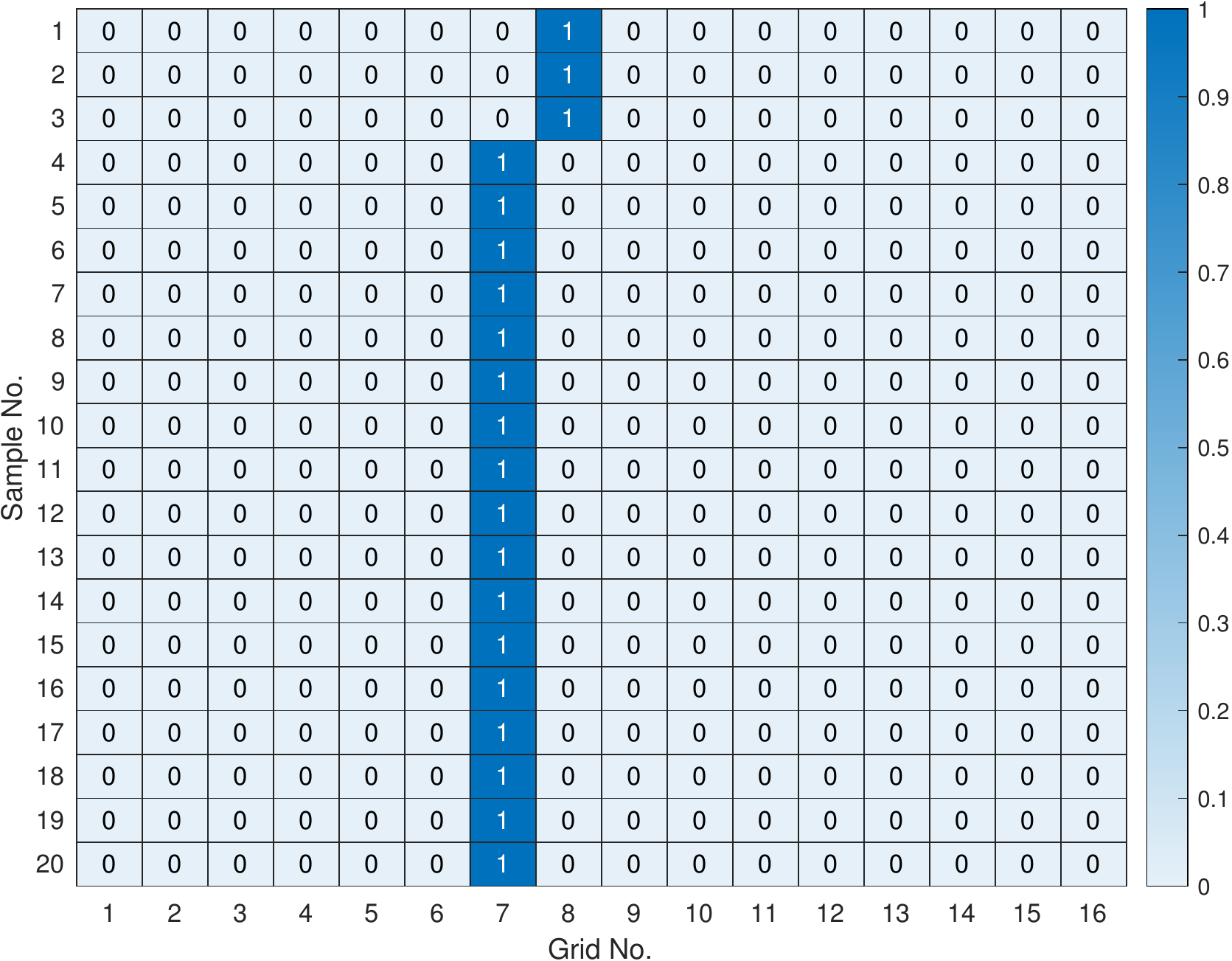}
\end{minipage}
}
\subfigure[]
{
\begin{minipage}[t]{0.45\linewidth}
\centering
\includegraphics[width=6cm]{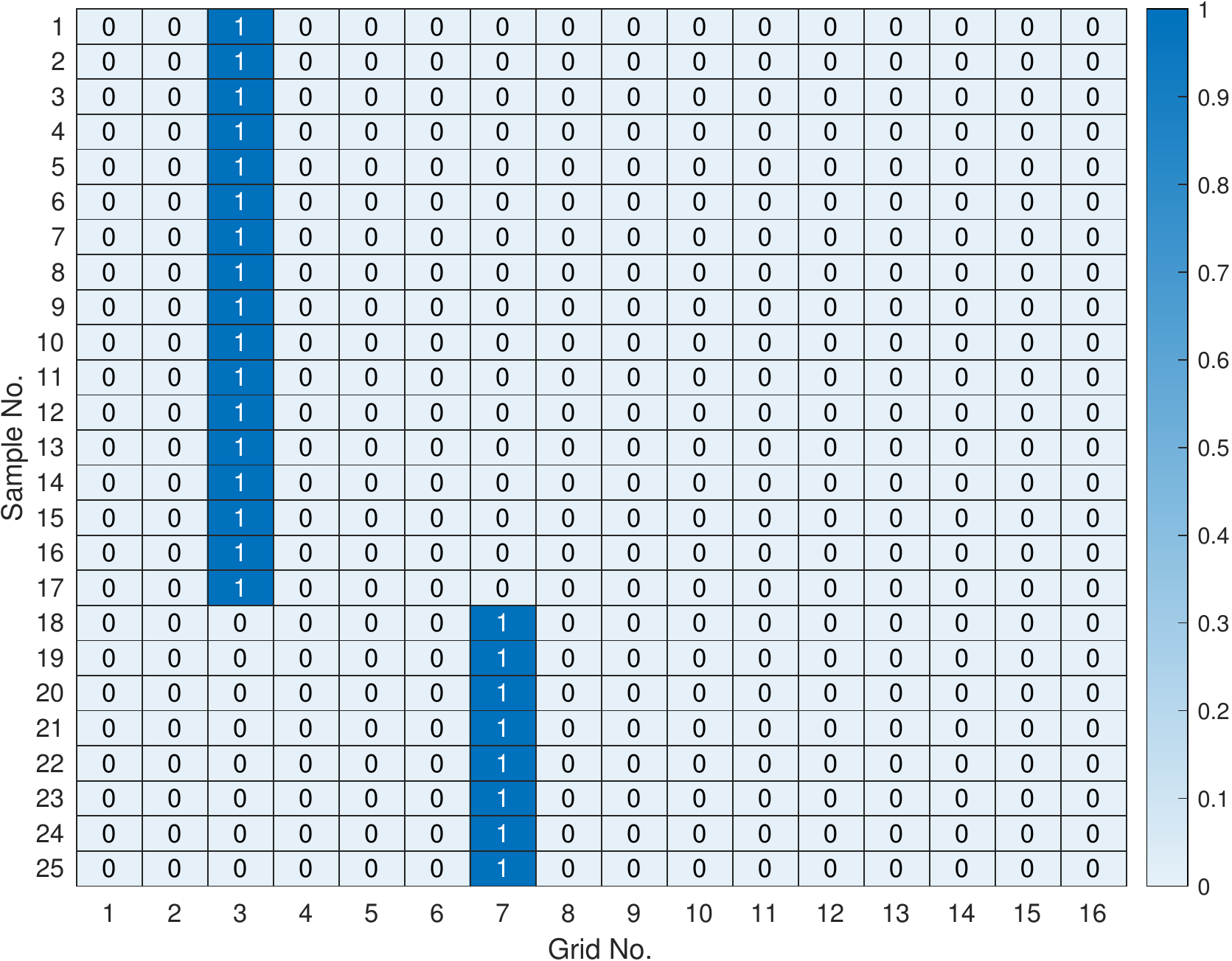}
\end{minipage}
}

\subfigure[]
{
\begin{minipage}[t]{0.45\linewidth}
\centering
\includegraphics[width=6cm]{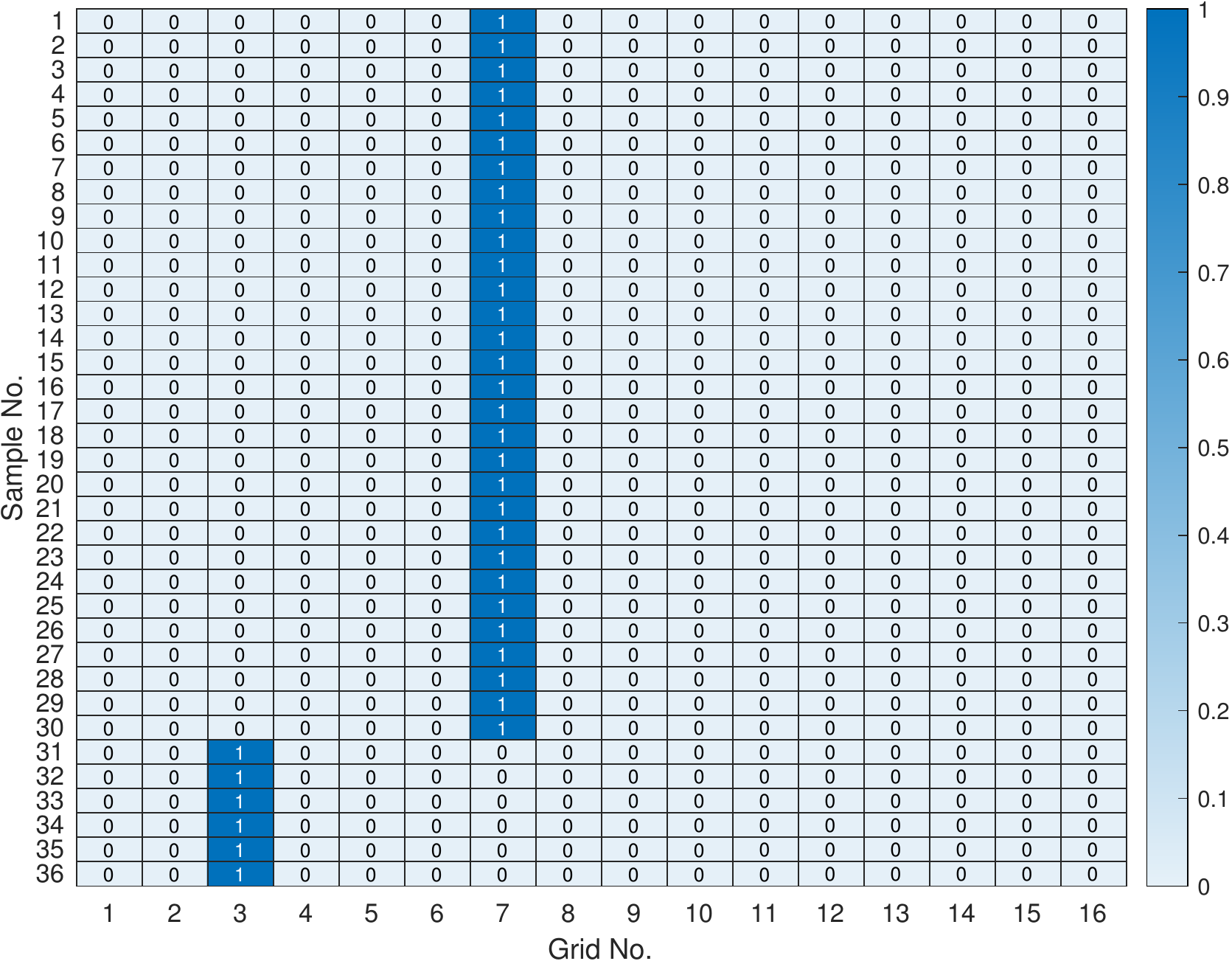}
\end{minipage}
}
\subfigure[]
{
\begin{minipage}[t]{0.45\linewidth}
\centering
\includegraphics[width=6cm]{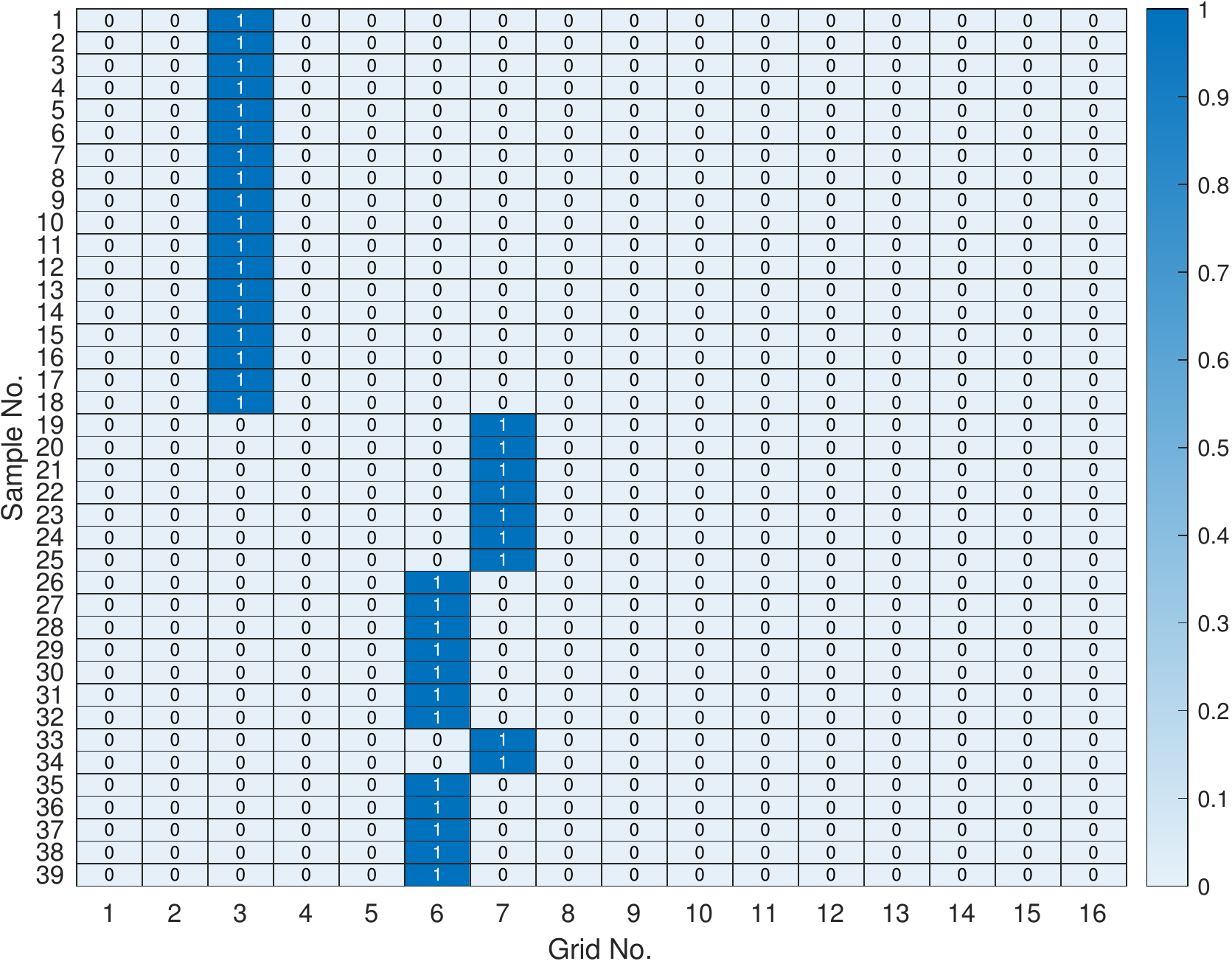}
\end{minipage}
}
\caption{Grid map representation of the vehicle trajectories shown in Fig. 5: (a) first trajectory, (b) second trajectory, (c) third trajectory, and (d) fourth trajectory (The blue grid indicates the visited location of the vehicle) when the number of grid $G$ is 16.}
\label{FIG7}
\end{figure*}

\subsubsection{Statistical Distribution of Vehicles}
\begin{itemize}
\item Step 5: Repeat Steps 1 through 4 for all vehicles and predict the next location of all vehicles as $\mathbf  {Y}_k=[{\hat y}_1, {\hat y}_2, \dots, {\hat y}_{V}]$, where $V$ is the number of vehicles at time $k$. $V$ is time-varying, as there are random vehicles that start and stop their trips at time $k$.

\item Step 6: According to the predicted location of each vehicle, we calculate the number of vehicles in each grid and denote the vector as $\hat G_k=[\hat g_1, \hat g_2, \dots, \hat g_i, \dots, \hat g_G]$, where $\hat g_i$ is the predicted number of vehicles in Grid $i$.
\end{itemize}

\begin{table}[!ht]
\renewcommand{\arraystretch}{1.2}
\centering
\caption{Network configurations of the proposed CapsNet-based vehicle trajectory prediction model for Porto city when the number of grid $G$ is 16.}
\scriptsize
\begin{tabular}{c c c}
\hline
\hline
\multirow{1}{*}{\textbf{Layers}} & \multirow{1}{*}{\textbf{Parameters}}  & {\textbf{Value}} \\
\hline
\multirow{3}{*}{Convolution layer} & {Filters}  & {80} \\
\cline{2-3}
& {Kernel size}  & {[1, 2]} \\
\cline{2-3}
& {Strides}  & {[1, 2]} \\
\hline
\multirow{4}{*}{Basic capsule layer} & {Dimensions}  & {4} \\
\cline{2-3}
& {Channels}    & {20} \\
\cline{2-3}
& {Kernel size} & {[1, 8]} \\
\cline{2-3}
& {Strides}  & {[1, 1]}  \\
\hline
\multirow{2}{*}{Advanced capsule layer} & {No. of capsules}  & {16} \\
\cline{2-3}
& {Dimensions}  & {8} \\
\hline
{Fully connectedly layer} & Neurons & 100 \\
\hline
{Output} & Neurons & 16 \\
\hline
\hline
\label{Table2}
\end{tabular}
\end{table}

\begin{figure*}[!htb]
\centering
\includegraphics[width = 14cm]{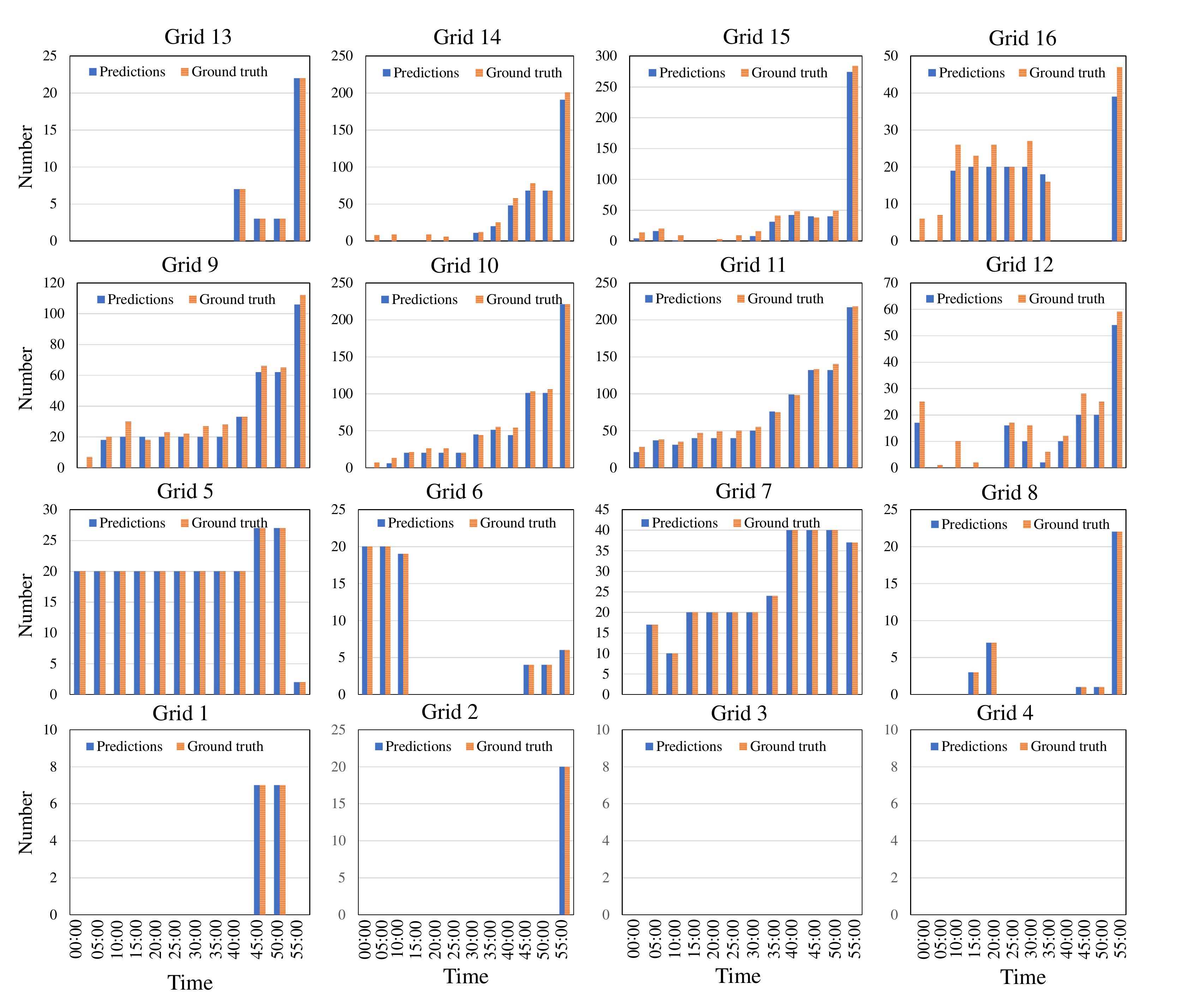}
\caption{Distribution of vehicles in each grid from time 16:40 to 17:40 on October 6 2014 when the number of grids $G$ is 16 (the red bar indicates the ground truth and the blue bar stands for predictions of the proposed method).}
\label{FIG8}
\end{figure*}

With time index $k$ grows, the distribution of vehicle trajectory could be timely updated Steps 1 through 6. If the number of vehicles quickly increases at the following time, it provides predictive decision support with offloading optimization for network sources allocation to ensure the quality of service.

\section{Result and Discussion}
This section evaluates the efficacy of the proposed method through real-world traces of taxi trajectories from two benchmarks for the fine-grained next location prediction, one of which is the data about the Porto city (Portugal) and the other dataset comes from Singapore. The first dataset provides a good benchmark, which can be used to compare the proposed method with state-of-art approaches. Although the data attributes of these two datasets are similar, the latter one is published recently and helps to evaluate the proposed method on large-scale data amounts with more vehicles and a high sampling rate.

\begin{table}[!ht]
\renewcommand{\arraystretch}{1.2}
\centering
\caption{Accuracy comparison in terms of index $P$ between the proposed method and the start-of-the-art approaches for Porto city when the number of grid $G$ is 16.}
\scriptsize
\begin{tabular}{c|c|c|c|c|c|c}
\hline
\hline
\multirow{2}{*}{$L$} & \multicolumn{4}{c|}{\textbf{Existing methods or models}} & \multirow{2}{*}{\textbf{Proposed}} & \multirow{2}{*}{IMA} \\
\cline{2-5}
& \textbf{LSTM$^{[22]}$} & \textbf{GRU} & \textbf{CNN} & \textbf{Transformer} & & \\
\hline
{3} & 89.35 & 94.83 & 94.51 & 94.77 & 95.84 & 7.26 \\
\hline
{7} & 89.82 & 95.40 & 94.90 & 95.13 & 95.91 & 7.78 \\
 \hline
{11} & 90.53 & 95.46 & 95.02 & 95.47 & 96.02 & 6.04 \\
\hline
{15} & 91.46 & 95.37 & 95.18 & 95.82 & 96.18 & 5.16 \\
\hline
{19} & 92.44 & 95.48 & 95.35 & 96.02 & 96.38 & 4.26 \\
\hline
\hline
\end{tabular}
\begin{tablenotes}
\item[1] IMA is short for improved prediction accuracy compared to LSTM.
\end{tablenotes}
\label{Table3}
\end{table}

\begin{figure}[!htb]
\centering
\includegraphics[width=7.5cm]{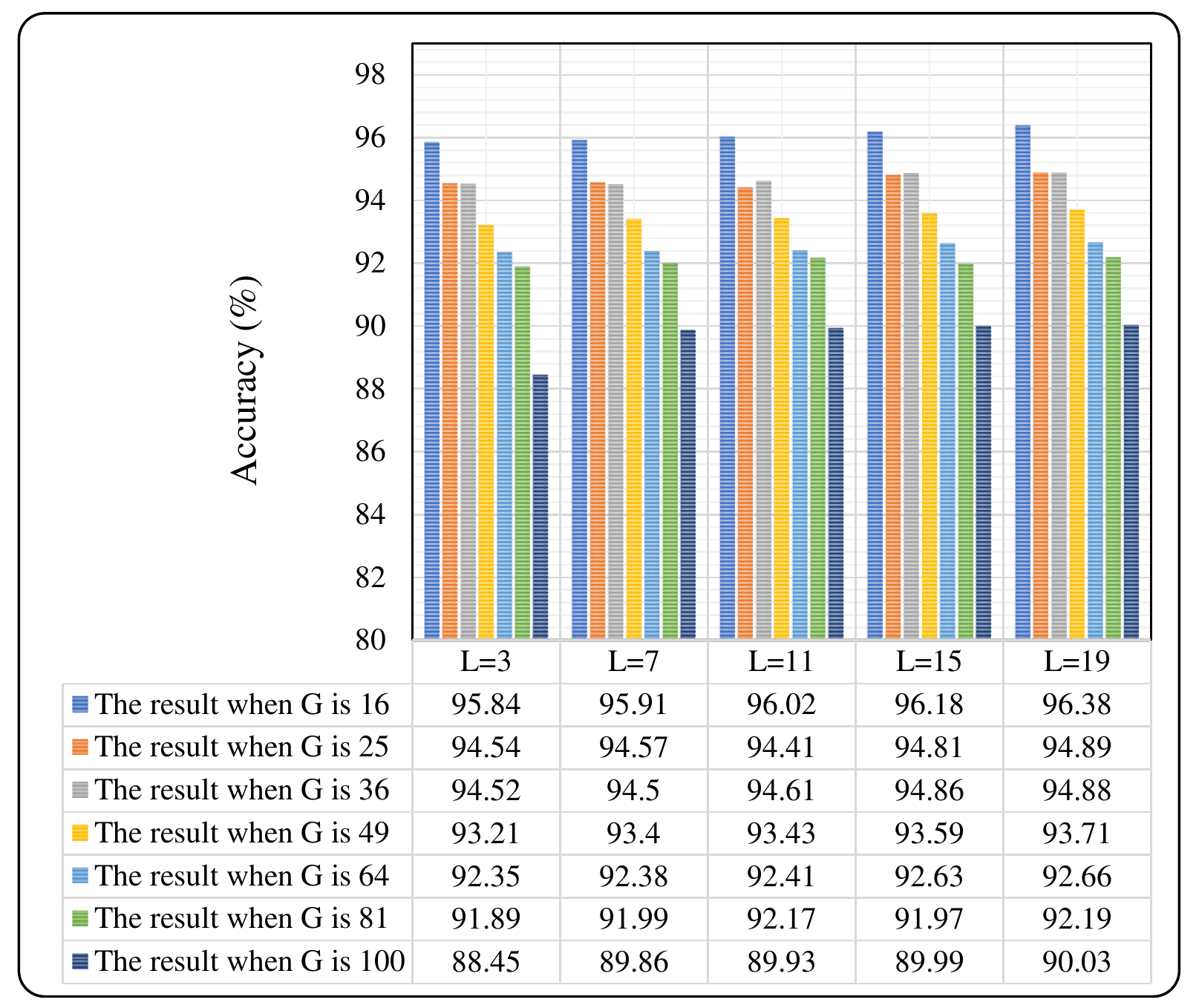}
\caption{Accuracy comparison of the proposed method in terms of $P$ under the different number of grids for the Porto city data.}
\label{FIG9}
\end{figure}

The metric prediction accuracy $P$ is defined as the ratio of correctly predicted grids over the total number of samples in the test dataset as follows,
\begin{equation}
P = \frac {1} {N_{T}} {\sum_{k} \phi (\mathbf{\Delta}(k) -\mathbf {\hat \Delta}(k))} \times 100\%
\end{equation}
where $\mathbf{\Delta}(k)$ is the true grid; $\mathbf{\hat \Delta}(k)$ is the predicted grid; $N_T$ is the number of testing samples; $\phi(\cdot)$ is equal to 1 if the true grid is same as the predicted grid; otherwise, $\phi(\cdot)$ will be 0.

\subsection{Results about Porto City Dataset}
442 taxis were recorded from July 2013 to December 2014 \cite{Ip2021} for analysis. Hundreds and thousands of trajectories are recorded in each vehicle through the in-vehicle installed GPS signal and then transmitted to data center for analysis. For each trajectory, recording of GPS signal starts at the beginning of the trip and ends when the trip stops. That is, the start/end information of this dataset is available. As the travel distance in each trip varies, the duration of each trajectory changes accordingly. The sampling interval keeps the same value of 15 seconds for all trips. For a better understanding of the moving trajectories of vehicles, Fig. \ref{FIG6} illustrates the first four trips of the first vehicle in the dataset.

\subsubsection{Data Representation with Grid Map}
By giving both values of $G_a$ and $G_o$ as 4, the total number of grids $G$ is 16, which follows the same configuration in \cite{Ip2021}. Correspondingly, the geographic information of Porto city has meshed into a defined grid of $16$. According to the geographic information of each grid, the vehicle trajectory at each sampling can be assigned to the corresponding grid. Fig. \ref{FIG7} exhibits the trajectory information in the grid map for vehicles illustrated in Fig. \ref{FIG6}.

Considering that the current location is closely related to its previous positions, we further slice the irregular trajectories with variable lengths into a series of trajectory frames. Specifically, five distinct lengths $L$ are determined as 3, 7, 11, 15, and 19 to follow the same configuration in \cite{Ip2021} for fair comparisons.

\subsubsection{CapsNet-based Prediction Model}
With grid map representation, the input of CapsNet can be arranged into an image format with the dimensions $16 \times L$. Correspondingly, the output of CapsNet is a vector with dimensions $16 \times 1$, each of which corresponds to a specific grid. The proposed model is configured according to the parameters listed in Table \ref{Table2}, which is used to examine the basic idea, and the best network structure is beyond the scope of this research. We have selected three typical machine learning models as baselines, including LSTM, GRU \cite{2014GRU}, CNN \cite{2015DNN}, and Transformer \cite{2017Transformer}.

\begin{itemize}
\item LSTM \cite{Ip2021}: It leverages the typical version of a recurrent neural network for trajectory prediction, consisting of an LSTM and a fully connected layer. The LSTM layer has 16 cells for sequential memorizing, and the fully connected layer with 16 neurons serves as the output.
\item GRU: We develop a prediction model with GRU, which requires less training data compared to LSTM. The designed network includes two layers. The GRU layer contains 16 cells for sequential memorizing, and the fully connected layer with 16 neurons serves as the output.
\item CNN: Treating the trajectory frame as image data, we have designed a CNN-based prediction model with one convolutional layer of 32 kernels and a max-pooling operation. The kernel size is selected as $(2,2)$.
\item Transformer: The self-attention mechanism is used to capture the hidden information at each time step. We have developed a Transformer-based prediction model with 16 heads \cite{2021Transformer}, which is the same as the number of grids.
\end{itemize}

It is worth mentioning that the results of LSTM is from the existing work \cite{Ip2021}. The results of GRU, CNN, and Transformer are gained by ourselves according to the aforementioned configurations. For all models, the data collected from July 2013 to July 2014 are employed as training data. And the testing data are collected from August 2014 to December 2014. 70$\%$ of training data are used for network training, and the remaining data are used for validation. The number of epochs is selected as 50 through early stopping.

Table \ref{Table3} compares the proposed CapsNet-based prediction with its counterparts in terms of the defined accuracy index $P$ using the testing data. We observe that the proposed method outperforms the LSTM-based model for each window length $L$. Besides, with the increase of $L$, the prediction accuracy improves correspondingly. The improved prediction accuracy (IMA) is calculated below,
\begin{equation}
IMA = \frac {P_{CpasNet}-P_{*}} {P_{*}} \times 100 \%
\end{equation}
where $P_{*}$ and $P_{CpasNet}$ stand for the estimation accuracy of a specific model and the proposed CapsNet, respectively.


\subsubsection{Statistical Distribution of Vehicles}
When the moving trajectory of each vehicle is predictable, it can deduce the distribution of all vehicles. Fig. \ref{FIG8} illustrates the predicted dynamic vehicle distribution from time 16:40 to 17:40 on October 6 in 2014. The number of vehicles every five minutes in each grid is calculated for analysis. Through analyzing the results, there are several new findings as follows:
\begin{itemize}
\item It is observed that the prediction meets well with the ground truth.
\item In each grid, the vehicle distribution varies largely over time. For instance, the number of visited vehicles increases significantly in Grids 9, 10, 11,  14, and 15 in an hour.
\item Vehicle distribution among grids differs from each other. By examining the results, distributions can be roughly classified into three categories: the increasing, then decreasing, and the steady trends. First, it is easy to know that the distribution in Grids 7, 9, 10, 11, 14, and 15 belongs to the increasing trend. Second, the vehicle distribution of Grid 6 meets with the decreasing trend. Third, the vehicle distribution of Grids 1, 2, 3, 4, 5, 8, 12, and 13 can be regarded as a steady trend.
\item With the help of predictive vehicle distribution, the operator can react to the dynamic demand of computation and communication sources in advance. To provide a high quality of service, we could deploy more base communication stations and edge computing devices for Grids 9, 10, 11, 14, and 15, where high commutation and computation demands are expected for the incoming vehicles. Contrarily, for grids with few vehicles, only a few road site units and edge computing devices will be enough to meet the requirement, avoiding deploying redundant devices to save costs.
\end{itemize}

\subsubsection{Robustness Analysis of $G$}
For vehicle trajectory prediction purposes, the space of a grid with the size of $500 \times 500 m^2$ is selected for a fair comparison with existing methods. In practice, the communication distance of edge computing and base station is limited and high dense deployment is required to ensure service quality. Correspondingly, a small size of grid is more suitable for mobility-aware services, such as offloading tasks. We further investigate the performance of the proposed method under different gird sizes. Specifically, the size of a space has been chosen as $400 \times 400 m^2$, $333 \times 333 m^2$, $286 \times 286 m^2$, $250 \times 250 m^2$, $222 \times 222 m^2$, and $200 \times 200 m^2$, corresponding to the value of $G$ as 25, 36, 49, 64, 81, and 100, respectively. Except for modifying the number of neurons in the output layer according to the specific value of $G$, the other layers follow the same network configuration as listed in Table \ref{Table2}. Fig. \ref{FIG9} illustrates the accuracy under each scenario, with consideration of historical information. The increase of $G$ challenges prediction accuracy, as the complexity increases. It is observed that the accuracy has been reduced by 6$\%$ from the $G=16$ to $G=100$ when the historical length $L$ is 19, but all results achieve the accuracy around 90$\%$.

\begin{figure}[!htb]
\centering
\includegraphics[width=7cm]{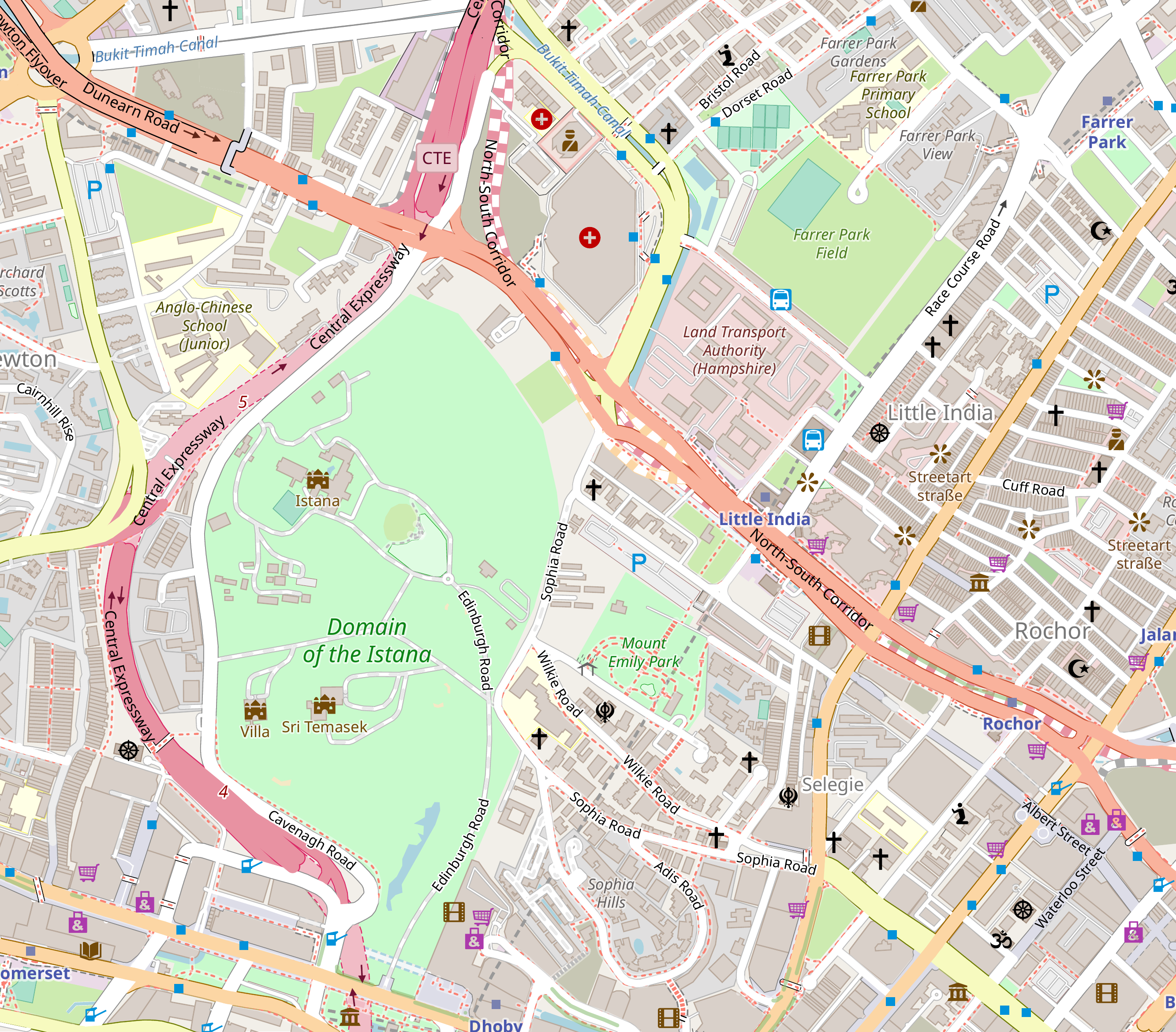}\\
\caption{The selected area in Singapore for vehicle trajectory prediction.}
\label{FIG10}
\end{figure}

\begin{table}[!ht]
\renewcommand{\arraystretch}{1.5}
\centering
\caption{Attributes of GPS data collected in Singapore \cite{Grab}.}
\scriptsize
\begin{tabular}{c c c}
\hline
\hline
\multirow{1}{*}{\textbf{Attribute}} & \multirow{1}{*}{\textbf{Data Type}}  & {\textbf{Description}} \\
\hline
ID & {String} & {Identifier for the trajectory} \\
\hline
Mode & String & Car or Motorcycle \\
\hline
Device & String & iOS or Android  \\
\hline
Latitude & {Float} & {World Geodetic System 1984} \\
\hline
Longitude & {Float} & {World Geodetic System 1984} \\
\hline
Timestamp & {Bigint}  & {Coordinated Universal Time}  \\
\hline
Accuracy Level & {Float}  & {Circle radius in meter} \\
\hline
{Bearing} & Float & Degrees relative to true north \\
\hline
{Speed} & Float & Measured by meters/second \\
\hline
\hline
\end{tabular}
\label{Table4}
\end{table}

\subsection{Results about Singapore Dataset}
A real-world GPS drivers' trajectory data about Singapore has been released by Grab company to describe the vehicle behaviors during two weeks from 08/04/2019 to 21/04/2019. In this dataset, the GPS trajectories are recorded by the driver's phone instead of in-vehicle devices, including Android and iOS devices, with a high-rate sample interval of one second. It should be pointed out that the sampling intervals in this dataset are not constant. Although most sampling intervals are around 1 second, sometimes the sampling rate can be as slow as 100 seconds. This challenges the prediction accuracy as the unequal sampling intervals.

To accelerate the algorithm implementation and verification, a local area surrounding Orchard Road in Singapore, which is the central business district and famous tourist attractions mostly visited, is selected for analysis. Fig. \ref{FIG10} shows the detailed spatial coverage of the concerned region in the dataset with an area of about 2km×2km. Considering the distance between Road side units is 200 meters, this specified area is divided into 100 city grids, in which there are 10 grids along the latitude and 10 grids along the longitude.

\begin{table}[!ht]
\renewcommand{\arraystretch}{1.2}
\centering
\caption{Network configurations of proposed CapsNet-based trajectory prediction model for the data collected in Singapore.}
\scriptsize
\begin{tabular}{c c c}
\hline
\hline
\multirow{1}{*}{\textbf{Layers}} & \multirow{1}{*}{\textbf{Parameters}}  & {\textbf{Value}} \\
\hline
\multirow{3}{*}{Convolution layer} & {Filters}  & {32} \\
\cline{2-3}
& {Kernal size}  & {[1, 100]} \\
\cline{2-3}
& {Strides}  & {[1, 1]} \\
\hline
\multirow{4}{*}{Basic capsule layer} & {Dimensions}  & {4} \\
\cline{2-3}
& {Channels}    & {8} \\
\cline{2-3}
& {Kernal size} & {[$L$, 1]} \\
\cline{2-3}
& {Strides}  & {[1, 1]}  \\
\hline
\multirow{2}{*}{Advanced capsule layer} & {No. of capsules}  & {32} \\
\cline{2-3}
& {Dimensions}  & {8} \\
\hline
{Fully connectedly layer} & Neurons & 200 \\
\hline
{Output} & Neurons & 100 \\
\hline
\hline
\end{tabular}
\label{Table5}
\end{table}

\begin{table}[!ht]
\renewcommand{\arraystretch}{1.2}
\centering
\caption{Accuracy comparison in terms of index $P$ between the proposed method and the start-of-the-art approaches for the data collected in Singapore.}
\scriptsize
\begin{tabular}{c|c|c|c|c|c|c}
\hline
\hline
\multirow{2}{*}{$L$} & \multicolumn{4}{c|}{\textbf{Existing methods or models}} & \multirow{2}{*}{\textbf{Proposed}} & \multirow{2}{*}{IMA} \\
\cline{2-5}
& \textbf{LSTM$^{[22]}$} & \textbf{GRU} & \textbf{CNN} & \textbf{Transformer} & & \\
\hline
{3}  & 85.65 & 84.51 & 77.52 & 80.11 & 93.32 & 7.67\\
\hline
{7}  & 84.62 & 82.69 & 77.13 & 79.26 & 91.66 & 7.04 \\
 \hline
{11} & 83.73 & 80.15 & 74.38 & 78.43 & 90.12 & 6.39 \\
\hline
{15} & 82.43 & 79.26 & 72.45 & 78.12 & 88.78 & 6.35 \\
\hline
{19} & 80.50 & 77.12 & 70.23 & 77.65 & 83.68 & 3.18 \\
\hline
\hline
\end{tabular}
\begin{tablenotes}
 \item[1] IMA is short for improved prediction accuracy compared to LSTM.
 \end{tablenotes}
\label{Table6}
\end{table}

\subsubsection{Data Representation with Grid Map}
The basic format and attributes of Grab data collected at each sampling interval are summarized in Table \ref{Table4}. The data are recorded from cars or motorcycles with installed Android or iOS devices. Here, only trajectories from Android devices installed vehicles are selected for analysis due to the high accuracy level. More than 4000 vehicles are observed in this data during two weeks. According to the data processing procedure, the raw trajectory data with latitude and longitude will be transformed into a series of temporal-spatial representations with several historical lengths, including 3, 7, 11, 15, and 19.

\subsubsection{CapsNet-based Prediction Model}
The first 70$\%$ data are used for training, and the remaining 30$\%$ data are used for testing. According to the given network configuration listed in Table \ref{Table5}, the proposed method yields the best accuracy compared to existing approaches: LSTM, GRU, CNN, and Transformer introduced in the previous dataset. The accuracy of the proposed method is more than 90$\%$, which defeats the existing approaches with an obvious improvement, as shown in Table \ref{Table6}.

\subsubsection{Statistical Distribution of Vehicles}
\begin{figure}[!htb]
\centering
\includegraphics[width=8cm]{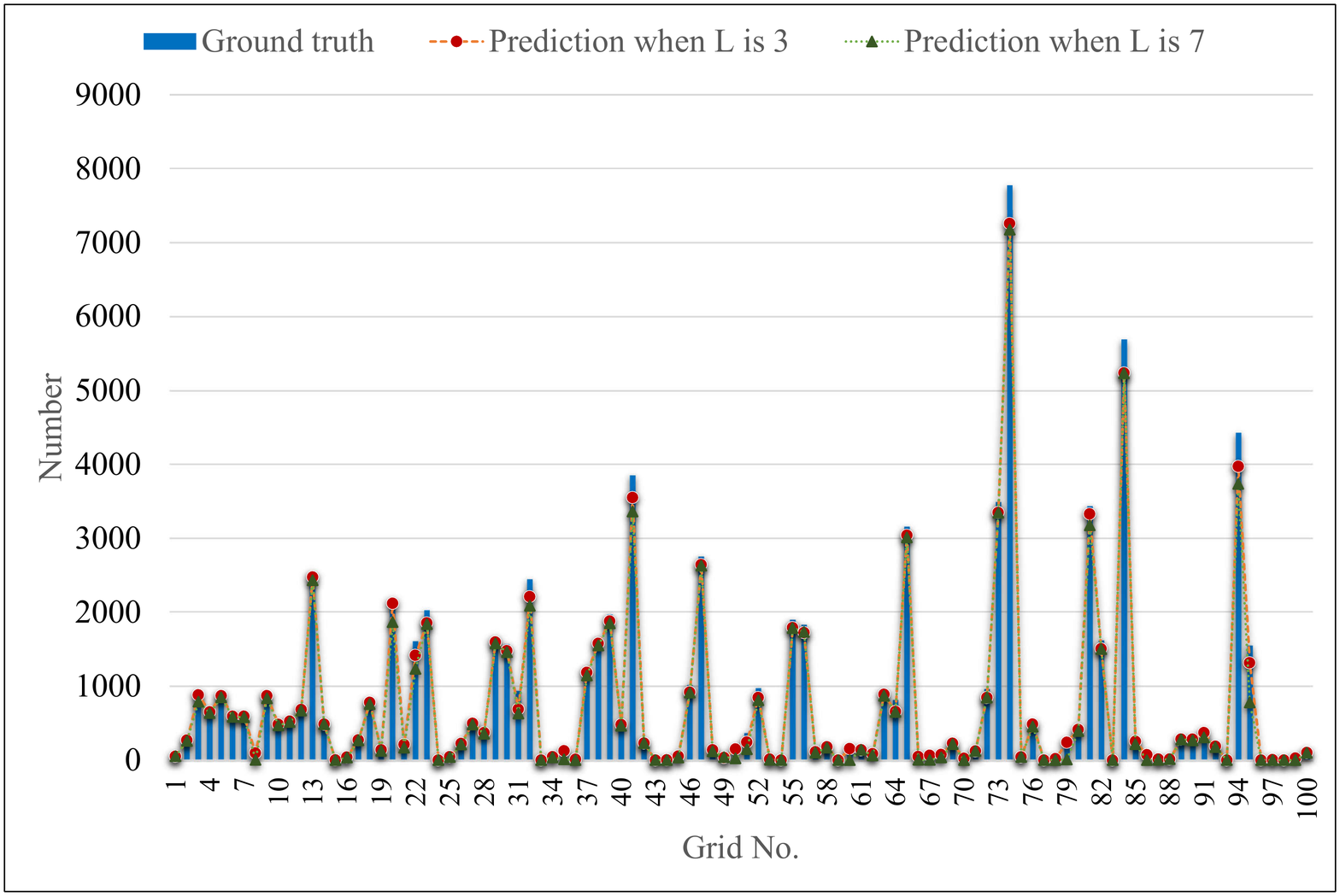}\\
\caption{Comparisons between the ground truth and predictions of vehicle distribution in the testing data on 17/4/2019 when the historical lengths $L$ are 3 and 7.}
\label{FIG11}
\end{figure}

\begin{figure}[!htb]
\centering
\includegraphics[width=8cm]{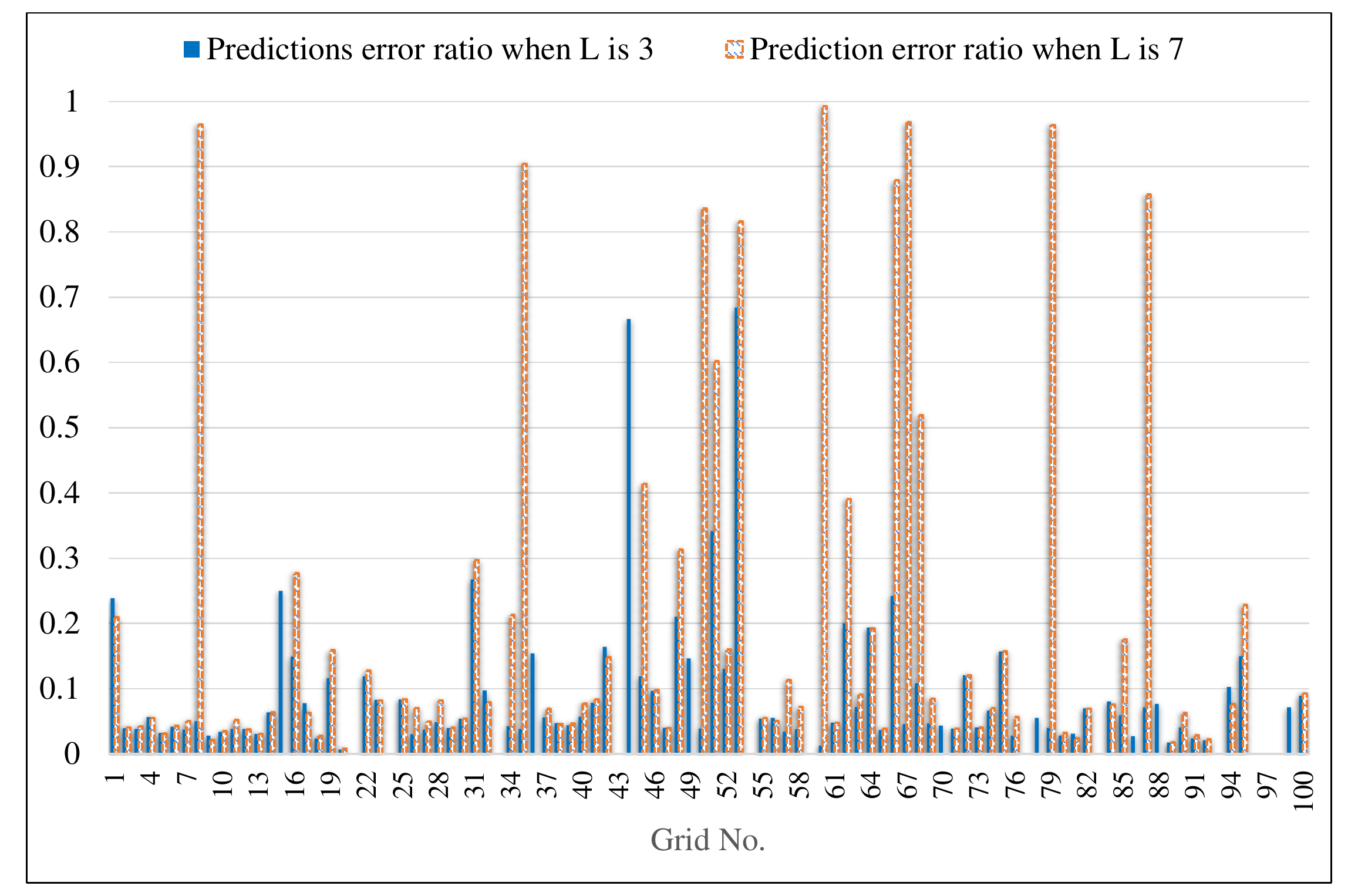}\\
\caption{Prediction error ratio of the proposed method when the historical length $L$ is 3 and 7.}
\label{FIG12}
\end{figure}

Time-varying distributions of vehicles in a specific grid can be inferred from the prediction results. Taking the historical length $L=3$ and $L=7$ as examples, Fig. \ref{FIG11} illustrates the predicted dynamic vehicle distribution on one day of 17/4/2019 in the employed grids, where the latitude is from 1.297348 to 1.315434 and the longitude is from 103.838447 to 103.856950. As vehicle data collected in Singapore are private hire cars, it is observed that the amounts of vehicles are small in some grids. When the historical length $L$ is 3, the number of predicted vehicles is very close to the ground truth after applying the well-trained prediction model for the testing data. Fig. \ref{FIG12} further demonstrates the prediction error ratio of each grid, and errors are less than 10$\%$ in most grids when $L$ is 3. The prediction error divided by its ground truth makes the prediction error ratio of a specific grid.

\textit{Remark:} The current results reveal the strength of the proposed method regarding vehicle trajectory prediction. In the application scenario of the V2X network, the benefits of this work still need more effort to integrate the prediction results with a specific topic. Among them, task offloading decision has received much attention, which is always formulated with a static and known number of users or tasks \cite{2020Offloading}\cite{V2X2}. Considering the high mobility of vehicles, the pre-defined number of users or tasks for edge computing will always be different from the actual value. Therefore, it is challenging to handle the mobility-aware computational efficiency-based task offloading and resource allocation scheme \cite{V2X3}.



All experiments are fairly conducted and compared under the same computing ability, which is achieved by a workstation equipped with eight processors of Intel Xeon E5-2620 v4 and a multi-graphics processor unit of NVIDIA GeForce GTX 1080Ti \cite{2021Qin}. Although computing complexity of the proposed method about training time will be large compared with its counterparts, including LSTM, CNN, and GRU, prediction time is almost the same for online computing when the model is trained.

\section{Conclusion and Future work}
To pursue a more reliable quality of service in the vehicle-to-everything (V2X) communication network, this paper proposed a capsule network-based model to predict future vehicle locations with its historical trajectory information. Capsule network makes full use of input data in a hierarchical structure, transmitting features from low-level to high-level in the capsule format. This work models the time-varying vehicle distribution with overall vehicle movements, timely informing the network operator to allocate network resources in advance. This purpose is achieved through a three-layer architecture, from the data representation with grid map, to vehicle trajectory prediction modeling based on capsule network, and online prediction of vehicle trajectory and vehicle distribution. Experiments conducted on the actual data from Porto city and Singapore have demonstrated the superiority of the proposed approach, which improves the prediction accuracy with different lengths of historical information. Moreover, the predicted vehicle distribution matches well with the actual case. Although an empirical way has been used to find the proper network structure of the capsule neural network, more efforts are deserved to explore automatic parameter configuration search for the capsule neural network.

Precise vehicle trajectory prediction paves the way to enhance various services in the V2X network. It is meaningful to further apply the prediction result to the resource allocation tasks, like the offloading decisions for communication or computation to the edge device, in which the number of users or tasks is always assumed to be constant. Therefore, joint consideration of dynamic changes in vehicle distribution will facilitate more efficient and practical resource allocation tasks.

\ifCLASSOPTIONcaptionsoff
\newpage
\fi


\begin{thebibliography}{1}
\bibitem{2022Zhang}
Y. Zhang, C. Li, T. H. Luan, C. Yuen, and Y. Fu, ``Collaborative driving: Learning-aided joint topology formulation and beamforming'', \emph{IEEE Vehicular Technology Magazine}, vol. 17, no. 2, pp. 103-111, 2022.

\bibitem{2020Liu}
W. Liu, Y. Watanabe, and Y. Shoji, ``Vehicle-assisted data delivery in smart city: A deep learning approach'', \emph{IEEE Transactions on Vehicular Technology}, vol. 69, no. 11, pp. 13849-13860, 2020.

\bibitem{Ref16}
M. Hasan, S. Mohan, T. Shimizu, and H. Lu, ``Securing vehicle-to-everything (V2X) communication platforms'', \emph{IEEE Transactions on Intelligent Vehicles}, vol. 5, no. 4, pp. 693-713, 2020.

\bibitem{Ref17}
B. Yang, X. Cao, K. Xiong, C. Yuen, and et al., ``Edge intelligence for autonomous driving in 6G wireless system: Design challenges and solutions'',  \emph{IEEE Wireless Communications}, vol. 28, no. 2, pp. 40-47, 2021.

\bibitem{RefSecurity1}
M. Li, L. Zhu, Z. Zhang, C. Lal, M. Conti, and F. Martinelli, ``Privacy for 5G-Supported Vehicular Networks'', \emph{IEEE Open Journal of the Communications Society}, vol. 2, pp. 1935-1956, 2021.

\bibitem{RefSecurity2}
M. Hasan, S. Mohan, T. Shimizu, and H. Lu, ``Securing Vehicle-to-Everything (V2X) Communication Platforms'', \emph{IEEE Transactions on Intelligent Vehicles}, vol. 5, no. 4, pp. 693-713, 2020.

\bibitem{SelfDrive}
K. Xiong, S. Leng, X. Chen, C. Huang, C. Yuen, and Y. L. Guan, ``Communication and computing resource optimization for connected autonomous driving'', \emph{IEEE Transactions on Vehicular Technology}, vol. 69, no. 11, pp. 12652-12663, 2020.

\bibitem{2020Offloading}
K. Xiong, S. Leng, C. Huang, C. Yuen, and Y. L. Guan, ``Intelligent task offloading for heterogeneous V2X communications'', \emph{IEEE Transactions on Intelligent Transportation Systems}, vol. 22, no. 4, pp. 2226-2238, 2021.

\bibitem{2021Offloading1}
Q. Ye, W. Shi, K. Qu, H. He, W. Zhuang, and X. S. Shen, ``Learning-based computing task offloading for autonomous driving: a load balancing perspective'', \emph{IEEE International Conference on Communications}, 2021, pp. 1-6.

\bibitem{2015Petit}
J. Petit, and S. E. Shladover. ``Potential cyberattacks on automated vehicles'', \emph{IEEE Transactions on Intelligent Transportation Systems}, vol. 16, no. 2, pp. 546-556, 2015.

\bibitem{Ref15}
B. Yang, X. Cao, C. Huang, C. Yuen, L. Qian and M. D. Renzo, ``Intelligent spectrum learning for wireless networks with reconfigurable intelligent surfaces'', \emph{IEEE Transactions on Vehicular Technology}, vol. 70, no. 4, pp. 3920-3925, 2021.

\bibitem{2014Zhu}
Y. Zhu, Y. Wu, and B. Li, ``Trajectory improves data delivery in urban vehicular networks'', \emph{IEEE Transactions on Parallel and Distributed Systems}, vol. 25, no. 4, pp. 1089-1100, 2014.

\bibitem{2021Irio1}
L. Irio, A. Ip, R. Oliveira, and M. Luis, ``An adaptive learning-based approach for vehicle mobility prediction,'' \emph{IEEE Access}, vol. 9, pp. 13671-13682, 2021.

\bibitem{1997LSTM}
S. Hochreiter, and J. Schmidhuber, ``Long short-term memory," \textit{Neural Computation}, vol. 9, no. 8, pp. 1735-1780, 1997.

\bibitem{Ref12}
S. Mozaffari, O. Y. Al-Jarrah, M. Dianati, P. Jennings, and A. Mouzakitis, ``Deep learning-based vehicle behavior prediction for autonomous driving applications: A review,” \emph{IEEE Transactions on Intelligent Transportation Systems}, pp. 1-15, 2020.

\bibitem{Ref13new}
X. Zhang, Y. Qin, C. Yuen, L. Jayasinghe, and X. Liu, ``Time-series regeneration with convolutional recurrent generative adversarial network for remaining useful life estimation," \emph{IEEE Transactions on Industrial Informatics}, vol. 17, no. 10, pp. 6820-6831, 2021.

\bibitem{Ref14new}
Y. Qin, S. Adams, and C. Yuen, ``A transfer learning-based state of charge estimation for lithium-ion battery at varying ambient temperatures," \emph{IEEE Transactions on Industrial Informatics},  vol. 17, no. 11, pp. 7304-7315, 2021.

\bibitem{RefZhouqi}
K. Q. Zhou, Y. Qin, B. P. L. Lau, C. Yuen, and S. Adams, ``Lithium-ion battery state of health estimation based on cycle synchronization using dynamic time warping," in \emph{IECON 2021-47th Annual Conference of the IEEE Industrial Electronics Society}, 2021, pp. 1-6.

\bibitem{2019Liu}
W. Liu, and Y. Shoji, ``Edge-assisted vehicle mobility prediction to support V2X communications,'' in \emph{IEEE Vehicular Networking Conference}, VNC, 2018, pp. 10227-10238.

\bibitem{2020Adege}
A. B. Adege, H. P. Lin, and L. C. Wang, ``Mobility predictions for IoT devices using gated recurrent unit network,'' \emph{IEEE Internet of Things Journal}, vol. 7, no. 1, pp. 505-517, 2020.

\bibitem{2021Irio}
L. Irio, and R. Oliveira, ``A comparative evaluation of probabilistic and deep learning approaches for vehicular trajectory prediction," \emph{IEEE Open Journal of Vehicular Technology}, vol. 2, pp. 140-150, 2021.

\bibitem{Ip2021}
A. Ip, L. Irio, and R. Oliveira, ``Vehicle trajectory prediction based on LSTM recurrent neural networks,'' in \emph{IEEE Vehicular Networking Conference}, 2021, pp. 1-5.

\bibitem{Choi2019}
S. Choi, J. Kim, and H. Yeo, ``Attention-based recurrent neural network for urban vehicle trajectory prediction," \emph{Procedia Computer Science}, vol. 151, 2019, pp. 327-334.

\bibitem{Liu2022}
J. Liu, Y. Luo, Z. Zhong, K. Li, H. Huang, H. Xiong, ``A probabilistic architecture of long-term vehicle trajectory prediction for autonomous driving," \emph{Engineering}, 2022, In press, DOI: doi.org/10.1016/j.eng.2021.12.020.

\bibitem{Izquierdo2022}
R. Izquierdo, Á. Quintanar, D.F. Llorca, et al. ``Vehicle trajectory prediction on highways using bird eye view representations and deep learning," \emph{Applied Intelligence}, 2022, In press, DOI: doi.org/10.1007/s10489-022-03961-y.

\bibitem{2020SubCapsNet}
M. Edraki, N. Rahnavard, and M. Shah, “Subspace capsule network” in \emph{Proceedings of the AAAI Conference on Artificial Intelligence}, vol. 34, no.7, 2020, New York, USA, pp. 10745-10753.

\bibitem{Ref11}
S. Sabour, N. Frosst, and G. E. Hinton,``Dynamic routing between capsules,'' in \emph{Proceedings of the 31th International Conference on Neural Information Processing Systems}, Long Beach, CA, USA, 2017, pp. 1-11.

\bibitem{CapsNet1}
R. P. Andr\'{e}s, L.D. Enrique, and P. Rodrigo, ``A novel deep capsule neural network for remaining useful life estimation,” \emph{Proc. Inst. Mech. Eng. O J. Risk Reliab.}, vol. 234, no. 1, pp. 151-167, 2020.

\bibitem{Grab}
X. Huang, Y. Yin, S. Lim, G. Wang, B. Hu, J. Varadarajan, and R. Zimmermann. ``Grab-Posisi: An extensive real-life GPS trajectory dataset in southeast asia,” in \emph{Proceedings of the 3rd ACM SIGSPATIAL International Workshop on Prediction of Human Mobility}, 2019, pp. 1-10.

\bibitem{Ref13}
Q. Liu, S. Wu, L. Wang, and T. Tan, ``Predicting the next location: A recurrent model with spatial and temporal contexts,” in \emph{Proceedings of the AAAI Conference on Artificial Intelligence}, 2016, pp. 194–200.

\bibitem{Ref14}
H. Liang, J. Wu, S. Mumtaz, J. Li, X. Lin, and M. Wen, ``MBID: Micro-blockchain-based geographical dynamic intrusion detection for V2X,” \emph{IEEE Communications Magazine}, vol. 57, no. 10, pp. 77-83, 2019.

\bibitem{TemporalCapsNet}
Y. Qin, C. Yuen, Y. Shao, B. Qin, and X. Li, ``Slow-varying dynamics-assisted temporal capsule network for machinery remaining useful life estimation," \emph{IEEE Transactions on Cybernetics}, vol. 53, no. 1, pp. 592-606, Jan. 2023,

\bibitem{2014GRU}
K. Cho, B. van Merriënboer, D. Bahdanau, and Y. Bengio, ``On the properties of neural machine translation: encoder-decoder approaches,'' in \emph{Proceedings of SSST-8, Eighth Workshop on Syntax, Semantics and Structure in Statistical Translation}, pp. 103-111, Doha, Qatar, 2014.

\bibitem{2015DNN}
Y. LeCun, Y. Bengio, and G. Hinton, ``Deep learning,'' \emph{Nature}, vol. 521, pp. 436-444, 2015.

\bibitem{2017Transformer}
A. Vaswani, N. Shazeer, N. Parmar, and et al, ``Attention is all you need," in \emph{Proceedings of the 31st International Conference on Neural Information Processing Systems}. Red Hook, NY, USA, 2017, pp. 1-15.

\bibitem{2021Transformer}
G. Zerveas, S. Jayaraman, D. Patel, A. Bhamidipaty, and C. Eickhoff. ``A transformer-based framework for multivariate time series representation learning," in \emph{Proceedings of the 27th ACM SIGKDD Conference on Knowledge Discovery $\&$ Data Mining}. New York, NY, USA, 2021, pp. 2114-2124.

\bibitem{2021Qin}
Y. Qin, W. T. Li, C. Yuen, W. Tushar, and T. Saha, ``IIoT-enabled health monitoring for integrated heat pump system using mixture slow feature analysis," \emph{IEEE Transactions on Industrial Informatics}, vol. 18, no. 7, pp. 4725-4736, 2022.

\bibitem{V2X2}
S. Raza, S. Wang, M. Ahmed, M. R. Anwar, M. A. Mirza, and W. U. Khan, ``Task offloading and resource allocation for IoV using 5G NR-V2X communication," \emph{IEEE Internet of Things Journal}, vol. 9, no. 13, pp. 10397-10410, 2022.

\bibitem{V2X3}
B. Li, F. Chen, Z. Peng, P. Hou, and H. Ding, ``Mobility-aware dynamic offloading strategy for C-V2X under multi-access edge computing," \emph{Physical Communication}, vol. 49, no. 101446, pp. 1-14, 2021.
\end{thebibliography}
\end{document}